\documentclass{article}

\usepackage{microtype}
\usepackage{graphicx}
\usepackage{subfigure}
\usepackage{amssymb}
\usepackage{amsmath}
\usepackage{tikz}
\usepackage{listings}
\usetikzlibrary{shapes,arrows,positioning}

\tikzstyle{process} = [rectangle, minimum width=1.5cm, minimum height=1.5cm, text centered, draw=black, fill=blue!10, font=\scriptsize]
\tikzstyle{arrow} = [thick,->,>=stealth]

\usepackage{booktabs} 

\usepackage{hyperref}

\usepackage[accepted]{icml2023}

\usepackage{amsmath}
\usepackage{amssymb}
\usepackage{mathtools}
\usepackage{amsthm}

\usepackage[capitalize,noabbrev]{cleveref}

\theoremstyle{plain}
\newtheorem{theorem}{Theorem}[section]

\theoremstyle{definition}

\theoremstyle{example}
\newtheorem{example}[theorem]{Example}

\theoremstyle{remark}

\usepackage[textsize=tiny]{todonotes}

\icmltitlerunning{Decision Optimization CoPilot}

\newcommand{\minimize}{\mbox{minimize\hspace{4mm} }}
\newcommand{\subto}{\mbox{subject to\hspace{4mm}}}
\newcommand{\boldunderline}[1]{\textbf{\underline{#1}}}

\begin{document}

\twocolumn[
\icmltitle{From Large Language Models and Optimization to Decision Optimization CoPilot: A Research Manifesto}



\icmlsetsymbol{equal}{*}

\begin{icmlauthorlist}
\icmlauthor{Segev Wasserkrug}{ibm}
\icmlauthor{Leonard Boussioux}{uwash,allen,lish}
\icmlauthor{Dick den Hertog}{uva}
\icmlauthor{Farzaneh Mirzazadeh}{mitibm}
\icmlauthor{Ilker Birbil}{uva}
\icmlauthor{Jannis Kurtz}{uva}
\icmlauthor{Donato Maragno}{uva}
\end{icmlauthorlist}

\icmlaffiliation{uwash}{Michael G. Foster School of Business, University of Washington, Seattle, WA, USA}
\icmlaffiliation{allen}{Allen School of Computer Science and Engineering, University of Washington, Seattle, WA, USA}
\icmlaffiliation{lish}{Laboratory for Innovation Science at Harvard, Cambridge, MA, USA}
\icmlaffiliation{ibm}{IBM Research - Israel}
\icmlaffiliation{uva}{Amsterdam Business School, University of Amsterdam, 1018TV Amsterdam, Netherlands}
\icmlaffiliation{mitibm}{MIT-IBM Watson AI Lab, IBM Research, Cambridge, MA}



\vskip 0.3in
]



\printAffiliationsAndNotice{} 

\begin{abstract}
Significantly simplifying the creation of optimization models for real-world business problems has long been a major goal in applying mathematical optimization more widely to important business and societal decisions.

The recent capabilities of Large Language Models (LLMs) present a timely opportunity to achieve this goal. Therefore, we propose research at the intersection of LLMs and optimization to create a \emph{Decision Optimization CoPilot} (\emph{DOCP}) - an AI tool designed to assist any decision maker, interacting in natural language to grasp the business problem, subsequently formulating and solving the corresponding optimization model.

This paper outlines our DOCP vision and identifies several fundamental requirements for its implementation. We describe the state of the art through both a literature survey and experiments using ChatGPT. We show that a) LLMs already provide substantial novel capabilities relevant to a DOCP, and b) major research challenges remain to be addressed. We also propose possible research directions to overcome these gaps.

We also see this work as a call to action to bring together the LLM and optimization communities to pursue our vision, thereby enabling much more widespread improved decision-making.
\end{abstract}

\section{Introduction}
\label{sec:introduction}

Many important real-world decision-making problems can be modeled and solved using mathematical optimization - a formal technique in which the problem is modeled by a set of decision variables subject to constraints, with the goal of optimizing some objective function. Mathematical optimization has provided huge financial and societal benefits in many cases, including more efficient food delivery by the United Nations  \cite{peters2022world}, millions of dollars in savings for e-commerce sites \cite{edelmanAlibaba}, and better radiotherapy treatment for cancer patients  \cite{edelmanRadiotherapy}. However, most decision-makers currently lack the necessary mathematical skills and/or access to the experts required to apply mathematical optimization. Consequently, it has been a long-standing goal to make optimization accessible to a broad range of decision-makers.

\emph{Large Language Models} (\emph{LLMs}) such as ChatGPT \cite{OpenAI2023ChatGPT} have democratized access to AI, making advanced AI skills accessible to almost anyone through natural language and significantly simplifying many tasks such as document and code generation. This technology has also resulted in the creation of a variety of intelligent \emph{copilots} that help users be much more effective and productive in their professional and personal lives (for example, the GitHub Copilot \cite{GitHubCopilot}  helps developers significantly accelerate software development through natural language interaction).

 Given the AI advancements enabled by LLMs, we believe the time is right to realize the vision of widely accessible optimization by bringing together LLMs with mathematical optimization. We propose the creation of a \emph{Decision Optimization CoPilot} (\emph{DOCP}): an AI at the disposal of every decision-maker that interacts with the user in natural language and utilizes the human's knowledge and feedback to create and solve a problem-specific optimization model, thereby leading to better decision making.

This article is organized as follows: 
after establishing the 
background in Section \ref{sec:background}, we detail our DOCP  vision and some 
required core capabilities in Section \ref{sec:DOCP}. In Section~\ref {sec:SOTA}, we summarize the state of the art and relevant gaps based on a literature survey and experiments with ChatGPT. We suggest research directions to address existing gaps to create a DOCP in Section \ref{sec:research-directions}.
Section \ref{sec:summary} ends with a summary and a call to action to realize our DOCP vision.  

\section{Background \label{sec:background}}
\subsection{Large Language Models\label{sec:LLM-background}} 

A \emph{language model} is mainly a model for conditional probability distributions over word sequences. 
\emph{LLM}s  
are deep neural network 
auto-regressive language models with billions or trillions of parameters, typically based on a specific architecture, \emph{transformers}, utilizing the \emph{attention mechanism} (see \cite{BahdanauCB14, vaswani2017attention}). In contrast to previous language models, which had to be handcrafted for specific tasks such as entity recognition or translation, LLMs are initially pre-trained on huge datasets, resulting in a general understanding of language captured in their large number of weights. They can then be adapted to a wide variety of tasks, including document generation, code generation, entity recognition, translation, and question-answering. There are now many LLMs, both 
open source 
(e.g. 
Flan-T5 \cite{chung2022scaling},
Llama~2 \cite{touvron2023llama2},
Mixtral \cite{jiang2024mixtral})
 and proprietary (GPT-3.5  with 175 billion  parameters
 \cite{fewshot_NEURIPS2020}, and GPT-4 \cite{openai2023gpt4} with  1.76 trillion parameters 
 underlying ChatGPT \cite{OpenAI2023ChatGPT}; and Gemini Pro with 137 billion parameters  underlying Bard \cite{Google_Bard_Gemini}).

Adapting an LLM to a specific downstream task can be done in a variety of ways, including:  \emph{fine-tuning} \cite{NIPS2015_7137debd,dodge2020fine,wei2022finetuned} - updating (some) of the network weights with additional, more domain-specific information, \emph{Reinforcement Learning with Human Feedback} (\emph{RLHF}) \cite{RLHF2019,NEURIPS2020_RLHF,NEURIPS2022_b1efde53}, in which a \emph{reward model} is trained by human feedback on the output of the LLM, and is then used to update the models weights, and \emph{prompting} \cite{fewshot_NEURIPS2020,sanh2021multitask,khot2022decomposed} - giving the LLM natural language instructions as part of the request. 


Some LLM applications include a workflow of steps external to the LLM, in some of which an LLM is used. Common workflow patterns of such LLM applications include \emph{Tree-of-thoughts} \cite{tot_NEURIPS2023,long2023large} and \emph{Graph of thought} \cite{GoT_besta2023}, both of which explore the solution space through a tree or graph like process, using backtracking and cycles when needed, thereby enabling a complex reasoning process mimicking human problem-solving patterns. 

\subsection{\label{sec:optim-background}Mathematical Optimization}
Mathematical optimization requires formally modeling a decision-making problems as:
\begin{equation}
\label{eq:optmodel}
\begin{array}{ll}
     \minimize & f(x) \\
     \subto & g(x) \leq 0, \\
     & h(x) = 0, 
\end{array}
\end{equation}
where $x\in \Re^n$ is the decision variables vector (representing the possible decisions),  $f(\cdot)$ is the objective function or goal to optimize, and the vector-valued functions $g(\cdot)$ and $h(\cdot)$ are used to define constraints on the acceptable solutions of the problem.

Once modeled as an optimization problem, there are many possible algorithms for finding a solution to the decision problem. The algorithms that can be used depend on the exact type of optimization model created. Efficient algorithms exist for \emph{Linear optimization} (\emph{LO}), which is appropriate when all functions in \eqref{eq:optmodel} are linear. \emph{Nonlinear optimization} extends this concept to situations where the objective or the constraint functions can be nonlinear functions. \emph{Mixed integer linear optimization} (\emph{MILO}) deals with problems of the form \eqref{eq:optmodel} where the objective and constraint functions are linear, but some or all decision variables are constrained to be integer. MILO is crucial in practical applications and is widely used in logistics, manufacturing, and supply chain management. However, unlike LO, most real-world MILO problems are NP-hard, making it computationally difficult to find exact solutions for large-scale real-world instances. Therefore, as there could be different optimization formulations for the same decision problem, an important modeling skill is finding the most efficient model. In addition, several advanced techniques have been developed to handle MILO problems efficiently. These strategies include, among others, decomposition or branch-and-bound methods, which break down complex problems into simpler subproblems, and cut generation techniques, which iteratively add constraints to simplify the solution search procedure \citep{wolsey1988integer}. As implementing these techniques is often problem-dependent, this also requires significant skill. 

Finally, due to the complexity of solving real-world problems using MILO, the use of \emph{optimization engines} such as Gurobi \citep{gurobi} and CPLEX \cite{cplex} has become common among practitioners. These engines receive a mixed integer linear optimization problem and solve the problem by employing the latest algorithms. In addition, they have many parameters that can provide additional efficiencies when fine-tuned for specific models. 


\subsection{Optimization Modeling}

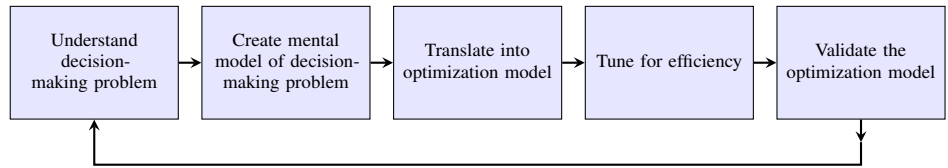
\begin{figure*}[h!]
    \centering
    \begin{tikzpicture}[node distance=0.3cm]
        \node (start) [process, text width=2cm] {Understand decision-making problem};
        \node (pro1) [process, right=of start, text width=2cm] {Create mental model of decision-making problem};
        \node (pro2) [process, right=of pro1, text width=2cm] {Translate into optimization model};
        \node (pro3) [process, right=of pro2, text width=2cm] {Tune for efficiency};
        \node (pro4) [process, right=of pro3, text width=2cm] {Validate the optimization model};
        \node (connect) [coordinate, below=of pro4] {}; 

        \draw [arrow] (start) -- (pro1);
        \draw [arrow] (pro1) -- (pro2);
        \draw [arrow] (pro2) -- (pro3);
        \draw [arrow] (pro3) -- (pro4);
        \draw [arrow] (pro4) -- (connect);
        \draw [arrow] (connect) -- ++(0,-0.3cm) -| (start.south); 
    \end{tikzpicture}
    \caption{Optimization Modeling Process}
    \label{fig:OptModelProcess}
\end{figure*}

Creating an optimization solution for a real-world business problem is a complex process requiring multiple steps.  This is because applying optimization requires a skilled optimization practitioner to work with the decision-maker using the process depicted in Figure \ref{fig:OptModelProcess}. 


As depicted in this process, the optimization practitioner needs to understand the problem and available data in sufficient detail and with enough precision to create a formal optimization model (in the form of Equation (\ref{eq:optmodel})). The practitioner then creates the model, attempting to find the most efficient formulation. The next step is often to apply advanced techniques such as problem decomposition, adding cuts and column generation, and correctly setting the parameter of the optimization engine so that the problem can be efficiently solved. Finally, the resulting optimization problem and solutions must be validated against the actual decision-making problem, with discrepancies resulting in additional iterations of this process. Moreover, all of these steps have to be carried out through constant interaction between the optimization modeler and the decision-maker, with the optimization modeler thinking in terms of mathematical formulations and the decision-maker communicating in terms of the problem domain. As a result, often multiple iterations of this process have to be carried out, 

\section{Decision Optimization CoPilot \label{sec:DOCP}}

We envision a \textbf{Decision Optimization CoPilot (DOCP)}: A friendly AI, incorporating an LLM, accessible to any business decision-maker who needs to make a decision that can benefit from optimization modeling. The DOCP will guide the decision-maker through deciding by interacting with the user in natural language and carrying out, behind the scenes, the process depicted in Figure \ref{fig:OptModelProcess}, ultimately resulting in an optimization model that provides good solutions to the specific business work. 

As a concrete example of such a decision problem, consider Example \ref{exa:Mobian} (based on a real-world use case [Anonymous 2023]). 

\begin{example}\label{exa:Mobian}
A company wants to have a methodology to identify optimal locations for their bike rental hubs. A hub is a parking facility where users (commuters and tourists) can park their cars, and bicycles are available for rental. The methodology should optimize the maximum coverage of the users' needs in the cities targeted to maximize the number of users and minimize the number of kilometers driven in the city.
\end{example}

The user would begin by providing the DOCP with the natural language description of the decision problem as in the example. The DOCP, through conversational interaction with the decision-maker, should first understand the precise objectives and constraints of the problem and the available data. Subsequently, the DOCP should validate that optimization is the right approach and which type of optimization model is required (e.g., LO model, MILO model, etc.) based on the problem characteristics (if optimization is not the right approach, the DOCP should notify the user and stop). The DOCP would then create the optimization model (possible models for the problem in Example \ref{exa:Mobian} appear in Appendix \ref{apx:mobian-optimization-models}) and run an appropriate algorithm (typically through an optimization engine) to solve it and provide actionable recommendations to the user. 

The DOCP would also need to be able to receive appropriate feedback from the user at any stage and course-correct or update the models and pipelines based on this feedback. Finally, the DOCP would need to communicate in a language understandable to the business decision-maker.

\subsection{Core Capabilities Required by a DOCP\label{seq:DOCP-requirements}}

\textbf{Requirement \#1 -  Translating a Business Level problem definition to an Optimization Model:} A  significant part of the optimization modeling process (Figure \ref{fig:OptModelProcess}) is the creation of a  \emph{mental model} of the decision-making problem at a sufficient level of detail and precision so that a formal optimization model can be created. Therefore, a DOCP would need to be able to guide the decision-maker through the modeling process that starts from (often a high-level) business problem definition and ends with the level of detail required for optimization model creation. Moreover, the DOCP would need to understand what data is available and decide what type of optimization model is required: Can it be modeled using an LO model, or does it require discrete decision variables, which would both require a MILO model (which would be much harder to solve)? Should uncertainty be considered, in which cases techniques such as Robust or Stochastic optimization \cite{Vladimirou1997} would be required? 

\textbf{Requirement \#2 -  Enabling the Decision Maker to Verify Optimization Model Correctness:} As LLMs can produce erroneous output, the optimization model generated by the DOCP could contain errors. While some generated model errors could manifest as runtime errors, there are often many much more subtle errors that would therefore be harder to detect. Such errors could include the generated objective not corresponding to the real-world decision objective or a missing or erroneous constraint. Using such erroneous formulations could result in the model being infeasible or generate solutions that, while optimal for the generated optimization problem, are not good solutions for the actual decision-making problem. When optimization experts are using LLMs,  the onus of validating the generated models can be placed on these experts. However, decision-makers who do not have optimization expertise cannot be expected to validate the model. Therefore, a DOCP must both a) take whatever steps are possible to reduce the number of errors in the model and b) provide sufficient tools appropriate for the business decision-maker not only to validate the model but also provide feedback in a way that will enable correction of the optimization model.

\textbf{Requirement \#3 - Creating Efficient Optimization Models:} Creating an optimization model that captures the real problem's constraints and objectives is insufficient.  As described in Section \ref{sec:optim-background} many times the resulting optimization problems are NP-hard. In such cases, many correct optimization models can still not provide useful solutions within a reasonable time frame.
Therefore, advanced modeling techniques such as finding equivalent, more efficient models, problem decomposition, column generation, and approximations must often be introduced. Moreover, when using off-the-shelf engines such as CPLEX (see section \ref{sec:optim-background}) to solve such models, optimization engine parameters such as cuts, branching variable selection, and primal/dual algorithms must be selected. As the most effective advanced modeling techniques or engine parameter values are often problem-specific, a DOCP must be able to apply these suitably to the specific problem. 

\section{State-of-the-Art with regards to DOCP Requirements\label{sec:SOTA}}

To understand the state-of-the-art capabilities relevant to a DOCP, we carried out a survey of related work and experimented with ChatGPT \cite{OpenAI2023ChatGPT}, a state-of-the-art LLM, as described below.

\subsection{\label{sec:related-work}Related work}

An increasing amount of work has leveraged LLMs
to convert natural language descriptions into mathematical models and formalizations \cite{zhang2019gap,meadows2022survey,WuJLRSJS22}. Of these, 
many works have used LLMs to generate optimization models with the goal of making optimization more widely accessible. The work of \citet{amarasinghe2023aicopilot} states as a goal the ability to ``leverage LLMs to support non-expert users to carry out business optimizations without having to consult experts". Focusing on production scheduling, this work constructed an open-source dataset consisting of relatively precise natural language problem descriptions and corresponding model formulations and used it to fine-tune a code-generating LLM to generate optimization models. This approach achieved good performance as measured by running the generated models, ensuring that they run correctly, and then comparing both execution time and obtained solution to the ``ground truth" models.  However, it does not address model validation by non-experts, nor explicitly how to generate efficient models.

OptiMUS \citep{ahmaditeshnizi2023optimus} uses an LLM to generate and code a mathematical model. It then uses an optimization engine to solve the model and run some feasibility tests to evaluate the correctness of the generated solutions. These unit tests can be automatically generated or suggested by optimization modeling experts. Despite showcasing the capability of integrating LLMs and optimization engines for automated optimization problem-solving, OptiMUS is contingent on a structured input representation (SNOP) and expert-created unit testing, presenting difficulties for users lacking proficiency in mathematical optimization.

 \citet{li2023synthesizing} propose a framework for automatic formulation from unstructured text descriptions of MILO problems. It utilizes a constraint classification scheme and constraint templates to guide the LLMs in creating MILO models from natural language descriptions. The authors assess the quality of the LLM by the quality of the generated models. In this work, the evaluation is entirely conducted by OR experts who must understand the correctness of the model.

There are additional works that have used LLMs to generate optimization models \cite{jang2022tag,
almonacid2023automatic,tsouros2023holy,
chen2023diagnosing}.  In 2022, the NL4Opt NeurIPS competition \cite{ramamonjison2023nl4opt} garnered significant attention in the realm of extracting linear optimization (LO) models from natural language. The competition was inspired by the pilot study of \citet{ramamonjison2022augmenting} where the authors introduced an automated formulation method named OptGen, designed to handle LO problem descriptions with diverse constraints. OptGen employs a two-step mapping strategy where, in the first step, all problem entities of the optimization problem (e.g., objective function, decision variables, etc.) are detected. In the second step, the problem description, together with the labeled semantic problem entities, is used to generate the final mathematical representation of the problem. For the entity detection problem,  \citet{ning2023novel} achieved the highest score, while  \citet{gangwar2023highlighting} achieved the highest accuracy in generating optimization models. This work utilized BART \cite{lewis-etal-2020-bart}, a large encoder-decoder model and enhanced the input by encapsulating entities with XML-like tagging. In all of these works, the quality of the solution was judged by how close the model was to the ``ground truth'' model.

Inspired by the NL4OPT challenge, \citet{tsouros2023holy} propose a modular framework for problem modeling via LLMs. The authors break down the modeling task into four subtasks, each implemented using prompting \cite{khot2022decomposed}. Beginning with a natural language description as input, the system sequentially extracts problem entities, establishes relationships, formalizes the optimization problem, and translates it into a constraint modeling language like CPMpy. Following the modeling phase, the code is compiled, executed, and automatically debugged if necessary. The system then engages in an interactive refinement process with the user, offering a comprehensive and efficient approach to problem modeling. In their survey paper, \citet{fan2024artificial} assessed LLMs' performance on textbook problems and real-world scenarios using the NL4OPT dataset and evaluation metrics. LLMs demonstrated impressive proficiency in modeling textbook-level problems; however, in the context of real-world problems, the generated models exhibited errors such as incorrect constraints, extra constraints, and missing constraints. Nonetheless, pointing out errors and prompting corrections proved effective, making LLM-generated formulations more accurate.

To summarize, a lot of work has focused on generating models from natural language descriptions. While good progress has been made, there are still many errors in the generated models. In addition, all works require quite a precise natural language description of the problem and not a high-level description, such as in Example \ref{exa:Mobian}, and only a few have gone beyond model generation to actually solve the generated models and analyze the results. Moreover, only \citet{ahmaditeshnizi2023optimus}  addresses the need to generate efficient models, and this also in a limited, implicit fashion (by having execution time as one of the measurement metrics). Finally, none of the above works address the need of a business user to verify the correctness of the model or the need to update the model as a result of business user feedback.

Unrelated to optimization, some works have attempted to use LLMs directly for complex reasoning tasks such as mathematical problem solving \cite{cobbe2021training,CoT_NEURIPS2022,lightman2023lets, trinh2024solving}.  Following this approach, \citet{yang2023large} employed LLMs to generate solutions for optimization problems, attempting to bypass the need for problem modeling and optimization engines. They applied this method to various Traveling Salesmen Problem (TSP) instances. In these instances, the LLM received node locations as input and produced a sequence as output. If the generated solution was not optimal, the authors provided the optimality gap as input and asked the model to refine its output iteratively. The approach adopted by the authors relies on the availability of an ``oracle'' that can generate the optimality gap and feasibility for a given solution.

However, previous works trying to use LLMs directly for reasoning (including \cite{bubeck2023sparks}) have shown that while the reasoning capabilities of LLMs are impressive, they still seem severely lacking. For example, in the aforementioned study of optimizing TSP problems, the performance of the LLM significantly deteriorates when dealing with large-sized problems. As the problem size increases beyond 10 nodes, the optimality gaps escalate rapidly, and optimization heuristics begin to surpass all LLMs. Furthermore, when a problem involves multiple constraints and nonlinear functions, the LLM is susceptible to generating solutions that are either infeasible or significantly distant from the optimum. In general, the work of \citet{gao2023pal} showed for that for multiple reasoning tasks, an LLM that first generates Python code and then runs that code to carry out the actual reasoning does much better than an LLM that directly tries to reason. This reinforces our proposed DOCP workflow: first generating an optimization model and then solving it externally to the LLM  to find good solutions.
 
There are also other works at the intersection of LLMs and optimization whose learning may be incorporated into a validation capability for the business user.  \citet{li2023large} uses an  LLM to provide a natural language interface to an existing optimization model. This interface can be used to query the model and its results and also provides some customization capabilities. \citet{chen2023diagnosing} develop a chatbot to assist an optimization expert in identifying and resolving infeasible optimization problems by taking Pyomo code as input.

\subsection{Experimentation Using ChatGPT}

We conducted experiments specifically targeting the three requirements described in Section \ref{seq:DOCP-requirements}. We did this using prompting with ChatGPT 4 \cite{OpenAI2023ChatGPT} as described below.

\subsubsection{Requirement \#1: From Business Level Problem Definition to Optimization Model\label{sec:req1-experiments}}

In this first set of experiments we played the role of a business decision maker, trying to work with ChatGPT to generate an accurate optimization model for the problem in Example \ref{exa:Mobian} through a natural language conversation. We provided ChatGPT with the appropriate context by creating personalized GPTs (see  \cite{OpenAI2023GPTs}) - versions of ChatGPT customized with natural language instructions.  We tried this with two GPTs with slightly different instructions, which explained to ChatGPT that its role is to have a conversation with the business user to understand the business problem, understand what type of optimization model is required, and subsequently create both a mathematical formulation of the problem and executable code suitable for running and solving this model using an optimization engine (the prompt for one of the GPT's appear in Figure \ref{fig:GPT-Context} and the full conversations in Appendix \ref{sec:exp-req1}).

\begin{figure}[t!]
\begin{center}
\centerline{\includegraphics[width=\columnwidth]{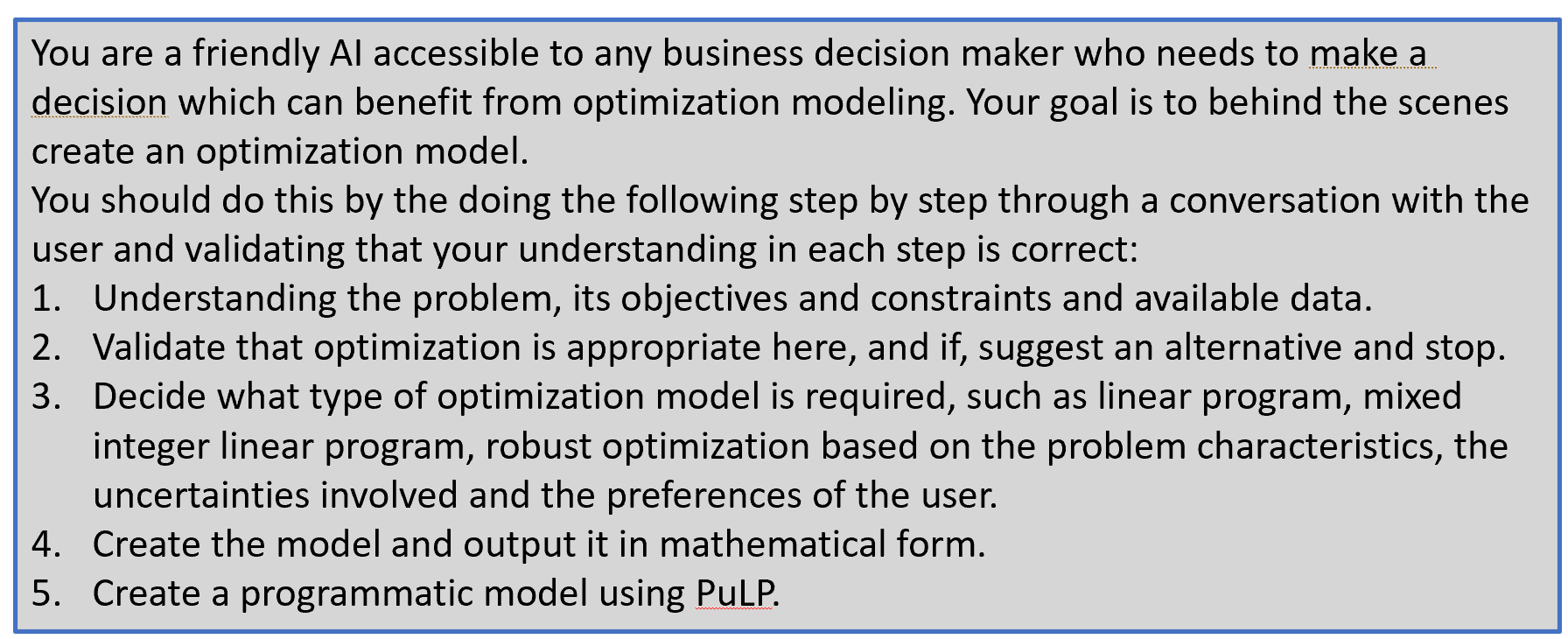}}
\caption{The Context Provided to the custom GPT}
\label{fig:GPT-Context}
\end{center}
\end{figure}

\begin{figure}[t!]
\begin{center}
\centerline{\includegraphics[width=\columnwidth]{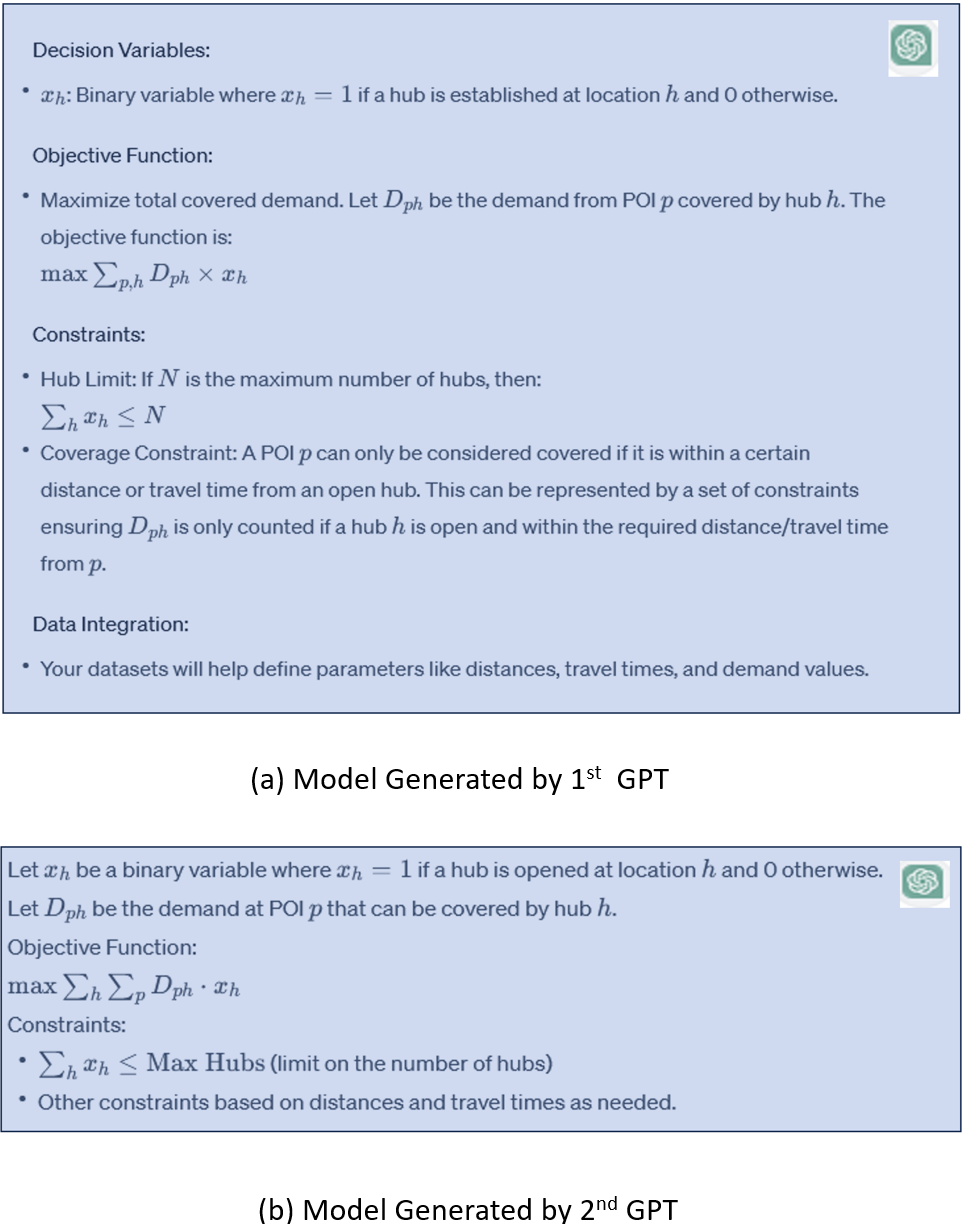}}
\caption{Optimization Models Generated by ChatGPT (Requirement \#1)}
\label{fig:GPT-Gap1Models}
\end{center}
\vskip -0.2in
\end{figure}

Both conversations began with a brief description of the problem as appearing in Example \ref{exa:Mobian}. In both cases, ChatGPT began correctly by asking for additional clarification and generated, in the end, a mathematical formulation of an optimization model. Moreover, in both cases, the optimization model selected by ChatGPT was a mixed integer linear optimization model, which is indeed appropriate for this problem. (The full conversations appear in Appendix \ref{sec:exp-req1}).

However, in both cases, the optimization problem formulation generated by ChatGPT (see Figure \ref{fig:GPT-Gap1Models}) was both incomplete and unsuitable for the business problem. The model was unsuitable as it incorrectly assigned a demand to a combination of a point of interest and a potential hub location (parameter $D_{p,h}$).  However, the demand assigned to a hub should be more granular, from a junction to a point of interest. Moreover, it made this decision even though the user mentioned explicitly that the available data is the demand from a junction to a point of interest. The model was also incomplete as it did not specify the constraints indicating whether a hub could cover a specific demand or not. While it asked for clarification to help it augment the constraints, it simultaneously asked the user to give feedback on the mathematical formulation. This is even though it was told in the context (Figure \ref{fig:GPT-Context}) that it was interacting with a business decision-maker. Finally, it did not directly generate the code but waited first for feedback on the mathematical formulation.

To summarize, the GPTs demonstrated impressive, relevant capabilities: they interacted with the user to understand the business problem and managed to get to the point of generating an optimization formulation. However, they did not take sufficient time to fully understand the problem before trying to generate a model, ending up with an unsuitable and incomplete formulation. Moreover,  after generating the mathematical formulation and before continuing to code generation, they asked the business user to confirm the correctness of the optimization model, which is not something a business user can do.

\subsubsection{Requirement \#2: Verifying Optimization Model Correctness}

\begin{figure}[t!]
\begin{center}
\centerline{\includegraphics[width=\columnwidth]{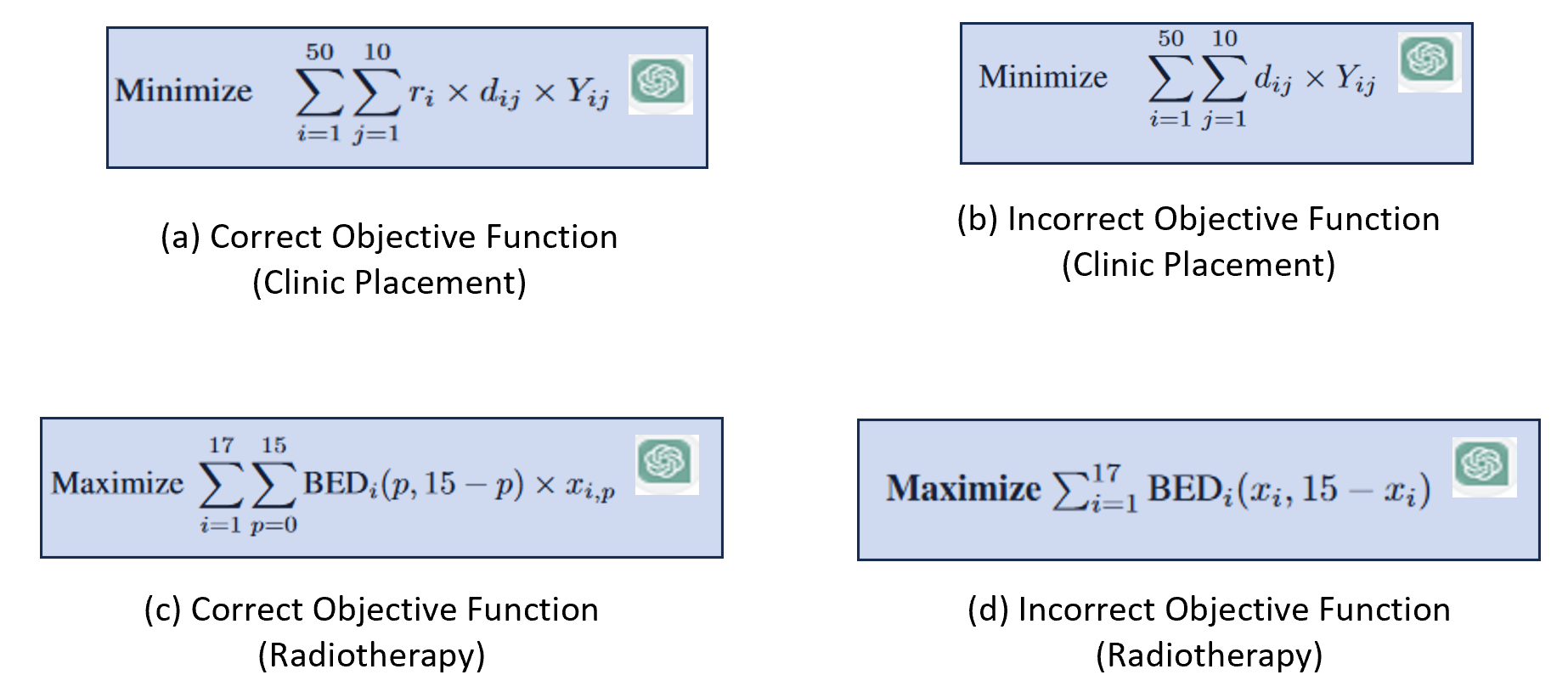}}
\caption{Objective Functions Generated by ChatGPT (Requirement 2)}
\label{fig:GPT-Req2Exp}
\end{center}
\vskip -0.2in
\vspace{-1.5em}
\end{figure}

In these experiments, we tested ChatGPT's ability to generate correct models for optimization problems when given a very precise description of the problem and not high-level ones as in Example \ref{exa:Mobian} (the full details of the experiments appear in Appendix \ref{sec:exp-req2}). The optimization problems were optimizing the location of vaccination clinics and optimizing radiotherapy treatment for cancer patients (a problem similar to the one in \cite{edelmanRadiotherapy}). Because ChatGPT's outputs are non-deterministic, we ran each ChatGPT five times on each problem. Overall, ChatGPT performed well, providing correct models for both problems four out of five times. However, for both problems, one of the five experiments generated an incorrect objective function expression (both correct and incorrect objective functions for both experiments appear in Figure \ref{fig:GPT-Req2Exp}). For the vaccination clinic optimization problem, it missed the essential term $r_i$ in the objective function, which is required to modify the demand by only assigning residents to clinics based on the area (see Figure \ref{fig:GPT-Req2Exp}(a) and \ref{fig:GPT-Req2Exp}(b) for the correct and incorrect objective function respectively). For radiotherapy treatment, one out of the five times, it incorrectly provided a nonlinear formulation of the problem, even though a linear formulation was explicitly asked for by incorporating the $BED$ nonlinear function (see Figure \ref{fig:GPT-Req2Exp}(d)). In the other four times, it correctly provided a linear objective function by calculating the values of the $BED$ function for the input data and summing over the values (Figure \ref{fig:GPT-Req2Exp}(c)). Both mistakes are relatively subtle and difficult for a novice optimization modeler to spot. This underscores the necessity for higher-quality model generation, validation mechanisms suitable for the business decision-maker, and the ability to incorporate user feedback to improve model accuracy. 

\subsubsection{Requirement \#3: Creating Efficient Optimization models}  
In this last set of experiments, our goal was to test whether ChatGPT can create models that are more efficiently solvable by optimization solvers.

We began by assessing ChatGPT's proficiency in transforming a correct formulation of the bike hub optimization model into a simpler one by prompting it to restructure the problem into an equivalent form with fewer variables. In this test, ChatGPT successfully produced an accurate version in only one out of five attempts. 

 In the four instances where the formulations were incorrect, ChatGPT exhibited a variety of errors. These included, among others, introducing nonlinear constraints or objective functions and incorporating constraints that only include parameters without variables (for the full experiment and list of errors, see Appendix \ref{apx:req3-exp1}). Furthermore, ChatGPT could not produce satisfactory reformulations without explicit instructions to maintain an equivalent form while minimizing variables. In scenarios with more open-ended instructions, ChatGPT struggled to deliver adequate solutions. It tended to oversimplify the problem, either by omitting constraints or by incorrectly redefining sets.

In another experiment, we offered ChatGPT a constraint with absolute values and asked ChatGPT if this constraint could be represented by linear constraint, which is much more efficient. For the case of a convex $\leq$ constraint, the answer was always correct. However, for the case of a concave $\geq$ constraint, ChatGPT did not use binary variables, and the answer was incorrect. We observed similar results for the case of `the sum of max of linear functions.' Such components often occur in, e.g., production-inventory problems, and it is well known that linear constraints can reformulate such a nonlinear constraint. ChatGPT also finds this reformulation but again forgets to introduce binary variables for the $\geq$ constraint.

Overall, this set of experiments shows that there is a significant gap in ChatGPT's ability to formulate efficient optimization models.

\subsection{Summary of State of the Art}

As can be seen, LLMs already provide important capabilities in realizing our DOCP vision. However, most of the progress of previous works has been in using precise and detailed natural language descriptions of a problem to generate correct optimization models (Requirement \#2). Little or no work has been done to generate optimization models from a business-level problem description or to ensure the creation of models that can be efficiently solved by solvers (Requirement \#1 and Requirement \#3), and our experiments show significant gaps in these areas. Even for requirement \#2, the current state of the art is such that often incorrect models are generated, including subtle errors that are very hard to detect. Therefore, much work remains for the DOCP vision to be realized.

\section{ Research Directions for Creating a DOCP\label{sec:research-directions}}

In creating a DOCP, LLM adaptation can play a significant part. As there are many ways in which an LLM can be adapted to a specific task (see Section \ref{sec:LLM-background}), an important research direction is how best to combine the different adaptation techniques - prompting, fine-tuning and RLHF. Specific examples include: should LLMs be fine-tuned with specific decision optimization knowledge? What can prompting techniques (such as \emph{Chain of thoughts prompting} \cite{CoT_NEURIPS2022} - a prompting technique found effective in reasoning tasks) accomplish? What are the best prompting techniques for each of the requirements? How should LLMs be adapted to generate correct and efficient optimization models? How best could LLMs be adapted to reduce the number of variables, reformulate nonlinear models into linear, and add problem-specific decomposition and column generation?

Another important research direction is which parts of the DOCP should be driven by the LLM, and which parts by external workflows such as Tree of Thoughts \cite{tot_NEURIPS2023} or Graph of Thoughts \cite{GoT_besta2023}? For a concrete example, it may be possible to utilize 
existing methodological frameworks, such as  \textit{CRISP-DM} (CRoss Industry Process for Data Mining \cite{Wirth2000CrispdmTA}) that guide an analytics practitioner through the process described by Figure \ref{fig:OptModelProcess}). Would a graph-of-thought DOCP implementation incorporating such methodologies yield good results?

Regarding Requirement \#2, as there is always a chance that an incorrect model will be generated, tools are required to enable the business user to validate any solution provided by a DOCP. Even today, business users often struggle to validate and understand solutions provided by expert-created optimization models, so tools such as what-if analysis are used. Specifically, business users can change a solution or provide an alternate one and see which model constraints are violated and what the resulting objective function value is. These existing techniques could be augmented by LLM capabilities such as those in \citet{li2023large}, where the user can ask questions about the model in natural language. It may also be worthwhile to explore creating LLM-based chatbots to whom the user can direct questions and have a conversation to understand better and validate the models.

Another important research direction for Requirement \#2 is how a specific generated model and the DOCP's general capabilities can be updated based on user feedback. For example, during the validation of a model, the user may realize that for specific solutions, the wrong objective has been calculated, or a business constraint has been wrongly marked as being violated. Similarly, the user may be able to provide solutions that satisfy the business constraints, possibly along with their objective function value. It would be important to both correct the specific model and continuously improve the DOCP based on this feedback.

Regarding requirement \#3, when an optimization engine is used to solve an optimization model, the parameter settings, such as which heuristics to use, can have a significant effect. A large body of existing works known as \emph{Learning to Optimize} \cite{LearntoOptimzeChen2022,cheng2022rlcg,gupta2022lookback,khalil2022mip} uses machine learning techniques to find good heuristic settings for specific problem types to reduce the solution time. Currently, such techniques are used by optimization experts, tailoring them to specific problem instances. An interesting avenue of research would be to see if LLMs could be adapted to select and customize the appropriate learning to optimize techniques in a problem-specific manner. As a final point, could the huge body of knowledge on the internet about the theory and application of optimization, including problem modeling and best practices (see, e.g., \citet{bookModelBuildingOptWilliams2013}), be better used to fine-tune an LLM to generate more efficient optimization models and optimization engine parameters' settings?

\subsection{Datasets and Benchmarks}

Research towards a DOCP will require appropriate datasets and well-defined quality metrics. Clearly, any such dataset should consist of a set of optimization problems, each containing a business-level description and a correct and efficient ``ground truth'' optimization model. Creating these parts of the datasets could utilize journals (e.g., \citet{IJAA}) and books \cite{bookModelBuildingOptWilliams2013} describing real-world case studies and their associated models. It may also be possible to extend datasets created by existing works bringing together LLMs and optimization such as \citet{amarasinghe2023aicopilot}. However, in addition to the natural language problem descriptions and ``ground truth'' models, as the DOCP is required to have a conversation with the business user to understand the problem and create an appropriate model, a component that can carry out the business user side of the conversation is also required. An interesting avenue of research is to see whether an LLM independent of the DOCP can be used to emulate the business user's side of the conversation, to enable DOCP development and testing.

Regarding evaluation,
measures used in most existing works using LLMs to generate optimization models are based on the syntactical accuracy of the generated model as compared to the ``ground truth'' model (see  Section  \ref{sec:related-work}). However, the goal of a DOCP is to generate good \textbf{solutions} to the business problem, and two very syntactically different models could model the same identical set of constraint-satisfying optimal solutions. Therefore, we posit that the metrics should directly relate to solution quality. While we leave a precise definition of these metrics to future work, we propose that these metrics should be based on measures such as whether the solutions generated by the DOCP satisfy the constraint of the ground truth model, how good they are in terms of ground truth objective value, and their execution time running time (possibly building on and extending the metrics defined by \citet{amarasinghe2023aicopilot}).
    
\section{Summary and call to action\label{sec:summary}}
We outlined our vision to combine LLMs and optimization modeling to create a DOCP to widely democratize the use of optimization for decision-making. We outlined three core requirements, discussed the current state of the art, and demonstrated the significant gaps to overcome. Due to these gaps and the huge positive implications of realizing this vision, we hope this work serves as a call for action to mobilize the LLM and optimization research communities around this goal.

We would also like to highlight that even intermediary techniques and tools created on the path to a DOCP will already significantly democratize the use of optimization by reducing the level of expertise required and enhancing accessibility. For instance, the improved automatic generation of efficient optimization models will empower novice modelers to achieve greater effectiveness. An optimization modeler could utilize a chatbot to facilitate discussions with business users (per requirement \#1), bridging the communication gap between business decision-makers and optimization experts. Introducing a natural language interface tailored for business use will increase the acceptance and application of these models in real-world scenarios. Such incremental advancements could already result in much more widespread data and model-driven decision-making.

\section{Statement of impact}

The vision of DOCP, if realized, has the potential to make decision-makers much more effective in their jobs and provide immense benefits by enabling optimization to be used across a vastly larger number of real-world decision-making problems, thereby leading to much more widespread better decision-making. As can be seen in the use cases referenced in the article, such optimization-driven decisions result in significant monetary and societal benefits such as better food utilization and healthcare.

\subsection*{Ethical implications}
In any optimization solution, there are dangers if the recommendations are followed blindly and poorly understood by the \textbf{decision-maker} whose responsibility is to make the decision. Introducing a DOCP does not and should not lessen the decision-maker's responsibility. When collaborating with an optimization modeler, there's the added benefit of a knowledgeable professional who can provide insights. This expert not only grasps the intricacies of the model but can also elucidate the underlying assumptions and its potential limitations for the business decision-maker. Therefore, using a DOCP underscores the need for features that allow the business decision-maker to understand the proposed solution and its limitations deeply. In high-stakes cases, the business decision-maker should still consult with an optimization modeler to review the optimization model created by the DOCP. In very high-risk cases, it may be better for the business decision-maker to forgo the direct use of a DOCP and instead engage with an optimization modeler, who may then leverage the DOCP to accelerate the creation of the optimization solution, ensuring a more informed and cautious approach to decision-making.

The need for extra care and validation is an additional key advantage of our proposed approach: having the LLM generate an optimization model rather than directly produce a solution. This approach is beneficial because it is considerably easier for an optimization modeler to review and understand the assumptions and limitations of a mathematical optimization model than a solution directly generated by an LLM. This step allows for a more transparent and reviewable decision-making process, ensuring that the decision-maker is fully informed of all aspects of the solution.

\newpage

\bibliography{LLMOpt}
\bibliographystyle{icml2023}

\newpage
\appendix
\onecolumn
\section{A note about optimization notation}

In this work, we use the terms `Linear Optimization (LO)' and `Mixed Integer Linear Optimization' instead of, respectively, `Linear Programming' and `Mixed Integer Linear Programming', which are terms the optimization community now attempts to avoid.

\section{Experiments with ChatGPT}

\subsection{Experiments for Requirement \#1\label{sec:exp-req1}}

\subsubsection{Conversation with GPT \#1\label{seq:exp-req1-conv1}}

\boldunderline{User Input}

I work for Mobian, a company in the Netherlands, and I want to have a methodology to identify optimal locations for their bike rental hubs. A hub is composed both of a parking facility, where users can park their cars, as well as bicycles available for rentals. The methodology should optimize the service while offering a maximum coverage of the needs of the commuters in the cities targeted. 

\boldunderline{GPT's Response}

To assist you with creating an optimization model for identifying optimal locations for bike rental hubs, let's start by understanding the problem in detail. This will involve clarifying the objectives, constraints, and the data you have available. Here are some initial questions to guide our discussion:

\begin{enumerate}
    \item \textbf{Objectives}:
    \begin{itemize}
        \item What is the primary goal for the placement of these hubs? Is it to maximize coverage, minimize cost, maximize usage, or a balance of these factors?
        \item Are there specific targets or key performance indicators (KPIs) that Mobian aims to achieve with these hubs?
    \end{itemize}
    \item \textbf{Constraints}:
    \begin{itemize}
        \item Are there limitations on the number of hubs that can be opened?
        \item Are there budget constraints for establishing these hubs?
        \item Are there any geographical or regulatory constraints that need to be considered (e.g., certain areas where hubs cannot be established)?
    \end{itemize}
    
    \item \textbf{Data Availability}:
    \begin{itemize}
        \item What data does Mobian currently have? This could include data on commuter patterns, existing transportation networks, demographic information, current bike usage statistics, etc.
        \item Is there information available on potential sites for these hubs and their costs?
     \end{itemize}
     \item \textbf{Cities Targeted}:
     \begin{itemize}
         \item Which specific cities or areas in the Netherlands are being considered for this project?
        \item Is there a priority list among these cities based on market research or strategic importance?
    \end{itemize}
\end{enumerate}

Understanding these aspects will help us determine if optimization modeling is appropriate for this task and, if so, what type of optimization model would be most suitable.

\boldunderline{User Input}

I want to maximize the covered demand, and have a bound on the number of opened hubs. I am first considering Amsterdam, and the data that I have is the following:
\begin{enumerate}
    \item Demand for POI $p$, coming from junction $j$
    \item Distance from junction $j$ to hub $h$ (meters)
    \item Distance from hub $h$ to POI $p$ (meters)
    \item Distance from junction $j$ to POI $p$ (meters)
    \item Car travel time from junction $j$ to hub $h$ (min)
    \item Car travel time from junction $j$ to POI $p$ (min)
    \item Bike travel time from hub $h$ to POI $p$ (min)
    \item Dataframe containing the names (Roads) and coordinates (Latitude and Longitude) of the different junctions
    \item Dataframe containing the names (Name) and coordinates (Latitude and Longitude) of the different hubs
    \item Dataframe containing the adress (Adress) and coordinates (Latitude and Longitude) of the different POIs
    \item Dataframe contraining the expected values for the different, individual parameters
    \item Geodataframe containing the road network accessible to bikers of Amsterdam 
    \item Geodataframe containing the road network accessible to cars of Amsterdam 
\end{enumerate}

\boldunderline{GPT's Response}

Thank you for providing detailed information about your data and objectives for Mobian's bike rental hub optimization in the Netherlands. Based on what you've described, we can proceed with formulating an optimization model. Here's a step-by-step approach:

\begin{enumerate}
    \item \textbf{Understanding the Problem}
    \begin{itemize}
        \item Objective: Maximize covered demand for bike rentals.
        \item Constraints: Bound on the number of hubs that can be opened.
    \end{itemize}
    \item \textbf{Validating Optimization Appropriateness}: 
    
    Given the goal to maximize demand coverage and the presence of constraints like the limit on the number of hubs, an optimization model is appropriate for this problem.
    
    \item \textbf{Deciding the Type of Optimization Model}

Given the nature of your problem, a \textbf{Mixed Integer Linear Program (MILO)} seems suitable. This is because:
    \begin{itemize}
        \item You have a binary decision to make (whether to open a hub at a particular location or not), which is an integer decision.
        \item The objective and constraints can likely be expressed as linear functions of your decision variables.
    \end{itemize}
    
    \item \textbf{Creating the Optimization Model}

    Decision Variables:
    \begin{itemize}
        \item $x_h$: Binary variable where $x_h=1$ if a hub is established at location $h$ and $0$ otherwise.
    \end{itemize}
    
    Objective Function:
    \begin{itemize}
     \item Maximize total covered demand. Let $D_{ph}$ be the demand from POI $p$ covered by hub $h$. The objective function is:

    $\max \sum_{p,h} D_{ph} \times x_h$
    
    \end{itemize}

    Constraints:
    \begin{itemize}
        \item Hub Limit: If $N$ is the maximum number of hubs, then:

        $\sum_h x_h \leq N$
    
        \item Coverage Constraint: A POI $p$ can only be considered covered if it is within a certain distance or travel time from an open hub. This can be represented by a set of constraints ensuring $D_{ph}$ is only counted if a hub $h$ is open and within the required distance travel time from pp.
    \end{itemize}

    Data Integration:

    \begin{itemize}
        \item Your datasets will help define parameters like distances, travel times, and demand values. 
    \end{itemize}

    \item \textbf{Programmatic Model Using PuLP}

Once the mathematical model is confirmed, we can translate it into a programmatic model using the PuLP library in Python. This will involve defining the decision variables, objective function, and constraints in code, and then using PuLP's solver to find the optimal solution.

Before moving on to the next step, it's important to confirm that the proposed model aligns with your understanding and requirements. Additionally, if there are specific coverage criteria for a POI to be considered "covered" by a hub, please specify those as well.

\end{enumerate}
\subsubsection{Conversation with GPT \#2}

\boldunderline{User Input}

I work for Mobian, a company in the Netherlands, and I want to have a methodology to identify optimal locations for their bike rental hubs. A hub is composed both of a parking facility, where users can park their cars, as well as bicycles available for rentals. The methodology should optimize the service while offering a maximum coverage of the needs of the commuters in the cities targeted. 

\boldunderline{GPT's Response}

    To develop an optimization model for identifying the optimal locations for Mobian bike rental hubs in the Netherlands, let's start by understanding your problem in more detail:

    \begin{enumerate}
        \item \textbf{Objectives}: The primary objective seems to be to optimize the placement of bike rental hubs to cover the maximum needs of commuters. This could be further detailed as:

        \begin{itemize}
            \item Maximizing coverage: Ensuring the hubs are accessible to the largest possible number of potential users.
            \item Service optimization: Ensuring efficient operations, such as adequate bikes and parking spaces at each hub.
        \end{itemize}
    \item \textbf{Constraints}: These might include:
    \begin{itemize}
        \item Budget constraints for setting up and maintaining hubs.Geographic or zoning limitations.
        \item Capacity constraints for each hub (like the number of bikes or parking spaces).
    \end{itemize}
    \item \textbf{Data Availability}: What kind of data do you have or can acquire? This might include:
        \begin{itemize}
            \item Geographic data of the cities.
            \item Population density and commuter patterns.
            \item Existing transportation infrastructure.
            \item Potential costs associated with setting up and maintaining hubs.
        \end{itemize}
\end{enumerate}

Do these points accurately capture your problem? And could you provide more details on the constraints and available data? This will help in ensuring that the optimization model is well-suited for your needs.

\boldunderline{User Input}
I want to maximize the covered demand, and have a bound on the number of opened hubs. I am first considering Amsterdam, and the data that I have is the following:
\begin{enumerate}
    \item Demand for POI $p$, coming from junction $j$
    \item Distance from junction $j$ to hub $h$ (meters)
    \item Distance from hub $h$ to POI $p$ (meters)
    \item Distance from junction $j$ to POI $p$ (meters)
    \item Car travel time from junction $j$ to hub $h$ (min)
    \item Car travel time from junction $j$ to POI $p$ (min)
    \item Bike travel time from hub $h$ to POI $p$ (min)
    \item Dataframe containing the names (Roads) and coordinates (Latitude and Longitude) of the different junctions
    \item Dataframe containing the names (Name) and coordinates (Latitude and Longitude) of the different hubs
    \item Dataframe containing the address (Address) and coordinates (Latitude and Longitude) of the different POIs
    \item Dataframe contraining the expected values for the different, individual parameters
    \item Geodataframe containing the road network accessible to bikers of Amsterdam 
    \item Geodataframe containing the road network accessible to cars of Amsterdam 
\end{enumerate}

\boldunderline{GPT's Response}

Based on the detailed data you have provided and your objectives, we can proceed to structure an optimization model for Mobian's bike rental hubs in the Netherlands. Here's a step-by-step approach:

\begin{enumerate}
    \item \textbf{Confirming Optimization Suitability}: Given your objective to maximize covered demand within a constraint on the number of opened hubs, an optimization approach is appropriate. It will allow you to find the best locations for the hubs considering the demand, distances, and travel times.

    \item \textbf{Model Type Decision}: For your case, a Mixed Integer Linear Program (MILO) seems appropriate. This is because:
        \begin{itemize}
            \item You have linear relationships (like distances and times).
            \item The decision to open a hub at a location is binary (either a hub is opened or it's not).
            \item You have a mix of continuous variables (distances, times) and integer variables (number of hubs).
        \end{itemize}

    \item \textbf{Model Structure}:
    \begin{itemize}
        \item \textbf{Decision Variables}:
        \begin{itemize}
            \item Binary variables indicating whether a hub is opened at a potential location.
            \item Other variables might include the number of bikes at each hub, though this seems secondary to your primary objective.
        \end{itemize}
        \item \textbf{Objective Function}:
        \begin{itemize}
            \item Maximize the total demand covered by the opened hubs. This will be a function of the demand at each POI and the accessibility of those POIs from the opened hubs.
        \end{itemize}
        \item \textbf{Constraints}:
        \begin{itemize}
            \item Limit on the total number of hubs that can be opened.
            \item Constraints ensuring that the demand is only considered covered if it's within a certain travel time or distance from an opened hub.
            \item Possibly, additional constraints based on your operational parameters.
        \end{itemize}
    \end{itemize}
    \item \textbf{Mathematical Formulation}:
    Let $x_h$ be a binary variable where $x_h=1$ if a hub is opened at location $h$ and $0$ otherwise. Let $D_{p,h}$ be the demand at POI $p$ that can be covered by hub $h$.

    Objective Function:

    $\max\sum_h \sum_p D_{p,h}\cdot x_h$
    
    Constraints:
    \begin{itemize}
        \item $\sum_h x_h \leq \text{MaxHubs}$ (limit on the number of hubs)
        \item Other constraints based on distances and travel times as needed.
    \end{itemize}
    
    \item \textbf{PuLP Implementation}:
    
    The next step would be to translate this model into a programmatic form using Python and the PuLP library. This involves defining the decision variables, objective function, and constraints in code, then solving the model using PuLP's solver.
\end{enumerate}

Before moving forward to the PuLP implementation, do you confirm that this model structure aligns with your needs? Additionally, if there are specific thresholds for travel times or distances that define whether a demand point is considered covered by a hub, please provide these details.

\subsection{Experiments for Requirement \#2\label{sec:exp-req2}}

\subsubsection{Vaccination Clinics Location Optimization\label{seq:opt-clinic-location}}

The state of Massachusetts knows the legendary talent of the students at MIT, and the Covid task force decided to call the Analytics for a Better World students for help. In case of a future pandemic, the state wants to be ready and plan for the deployment of mass vaccination clinics in Eastern Mass.

Given available resources, you can build only 4 sites across the Boston area. You have identified 10 potential sites, each with a fixed capacity (i.e., a maximum number of residents that can be vaccinated there). All residents will need to access one of the clinics you will construct. You aim to determine which sites to build to optimize the vaccination service provided to the population.

The area under consideration can be modeled as a $5 \times 10$ miles rectangle, with 50 major residential areas. We index the residential areas by $i = 1, \ldots, 50$ and the vaccination sites by $j = 1, \ldots, 10$. We use the so-called ``Manhattan distance'': if area $i$’s coordinates are $(x_{i}, y_{i})$ and site $j$’s coordinates are in $(x_{j}, y_{j})$ , the distance from $i$ to $j$ is given by $d_{ij} = |x_{i} - x_{j}| + |y_{i} - y_{j}|$.

You are provided with the following data:

\textbf{Parameters}

\begin{itemize}
    \item $r_{i}:$ number of residents in area $i=1, \ldots, 50$;
    \item $C_{j}:$ capacity of vaccination site $j=1, \ldots, 10$;
    \item $d_{i j}:$ distance between area $i=1, \ldots, 50$ and site $j=1, \ldots, 10$.
\end{itemize}

You also have access to the following files:

\begin{itemize}
    \item P1\textunderscore areas.csv: A matrix of size $50 \times 3$ that indicates, for each area, (i) its x-coordinate, in miles (0 to 5), (ii) its y-coordinate, in miles (0 to 10) and (iii) its number of residents.
    \item P1\textunderscore clinics.csv: A matrix of size 10×3 that indicates, for each candidate clinic, (i) its x-coordinate, in miles (0 to 5), (ii) its y-coordinate, in miles (0 to 10), and (iii) its capacity.
\end{itemize}

Formulate an integer optimization model that optimizes the selection of sites to minimize the total distance across all residents required to access their assigned clinic.

\boldunderline{Correct ChatGPT's answer}

To formulate an integer optimization model for selecting vaccination sites that minimizes the total distance for all residents to access their assigned clinic, we define the decision variables, objective function, and constraints as follows:

\textbf{Decision Variables}
\begin{itemize}
    \item Site Selection Variable ($y_j$): A binary variable for each potential vaccination site $j=1, \ldots, 10$, where $y_j=1$ if site $j$ is selected and $0$ otherwise.
    \item Assignment Variable ($x_{ij}$): A binary variable for each pair of residential area $i$ and vaccination site $j$, where $x_{ij}=1$ if area $i$ is assigned to site $j$ and $0$ otherwise.
\end{itemize}

\textbf{Objective Function}
\begin{equation*}
    \text{Minimize } \sum_{i=1}^{50} \sum_{j=1}^{10} r_i \cdot d_{ij} \cdot x_{ij}
\end{equation*}
where $r_i$ is the number of residents in area $i$, and $d_{ij}$ is the distance between area $i$ and site $j$.

\textbf{Constraints}
\begin{enumerate}
    \item Site Selection Limit: Only 4 sites can be selected.
    \begin{equation*}
        \sum_{j=1}^{10} y_j = 4
    \end{equation*}
   
    \item Assignment Constraint: Each area must be assigned to exactly one clinic.
    \begin{equation*}
        \sum_{j=1}^{10} x_{ij} = 1 \quad \text{for all } i=1, \ldots, 50
    \end{equation*}

    \item Capacity Constraint: The number of residents assigned to a clinic cannot exceed its capacity.
    \begin{equation*}
        \sum_{i=1}^{50} r_i \cdot x_{ij} \leq C_j \quad \text{for all } j=1, \ldots, 10
    \end{equation*}
    where $C_j$ is the capacity of clinic $j$.

    \item Assignment Only to Selected Sites: Residents can only be assigned to selected sites.
    \begin{equation*}
        x_{ij} \leq y_j \quad \text{for all } i=1, \ldots, 50, \text{ and } j=1, \ldots, 10
    \end{equation*}
\end{enumerate}

\textbf{Binary Nature of Decision Variables}
Both $x_{ij}$ and $y_j$ are binary variables, taking values in $\{0,1\}$.

\textbf{Model Summary}
\begin{itemize}
    \item \textbf{Objective}: Minimize the total distance traveled by all residents to their assigned vaccination clinics.
    \item \textbf{Constraints}:
    \begin{enumerate}
        \item Site Selection: Exactly 4 sites should be chosen.
        \item Assignment: Each residential area should be assigned to exactly one clinic.
        \item Capacity: The total residents assigned to a clinic should not exceed its capacity.
        \item Assignment Validity: Residents can only be assigned to selected clinics.
        \item Binary Variables: Decision variables are binary.
    \end{enumerate}
\end{itemize}

This integer optimization model effectively captures the requirements for the optimal deployment of vaccination clinics in the Boston area, aiming to minimize the total distance residents need to travel to get vaccinated while considering the constraints of clinic capacity and site selection.

\paragraph{Comments on the formulation}

The model presented by ChatGPT assumes that all residents in a given area are assigned to the same vaccination clinic. This assumption eases modeling and interpretation and is fine by the question formulation, as no constraint was specified. However, it is also possible to consider formulating the model at the individual resident level rather than aggregating residents by area. ChatGPT was able to carry correct extensions when prompted.

\subsubsection{Wrong formulation from ChatGPT}

\paragraph{Correction to the Objective Function of the Integer Optimization Model}
The following objective function in the integer optimization model from ChatGPT is incorrect as it fails to include the term \( r_i \), representing the number of residents in area \( i \). The correct objective function should weigh the distance \( d_{ij} \) by the number of residents \( r_i \) traveling to the site \( j \). Thus, the correct objective function to minimize the total distance traveled by all residents should be:
\[
\text{Minimize} \quad \sum_{i=1}^{50} \sum_{j=1}^{10} r_i \times d_{ij} \times Y_{ij}
\]
This modification is crucial as it ensures that the model accurately reflects the total distance traveled by the entire population rather than merely considering the distances between areas and clinics without regard to the number of residents in each area.

\paragraph{Answer from ChatGPT}

To formulate an integer optimization model for the problem of selecting optimal sites for vaccination clinics in the Boston area, we consider decision variables, an objective function, and constraints.

\paragraph{Decision Variables}
\begin{itemize}
    \item Binary Site Selection Variable: \( X_j \) where \( j = 1, \ldots, 10 \). \( X_j = 1 \) if site \( j \) is selected, and \( X_j = 0 \) otherwise.
    \item Assignment Variable: \( Y_{ij} \) where \( i = 1, \ldots, 50 \) and \( j = 1, \ldots, 10 \). \( Y_{ij} = 1 \) if residents of area \( i \) are assigned to site \( j \), and \( Y_{ij} = 0 \) otherwise.
\end{itemize}

\paragraph{Objective Function}
Minimize the total distance traveled by all residents to their assigned clinics:
\[
\text{Minimize} \quad \sum_{i=1}^{50} \sum_{j=1}^{10} d_{ij} \times Y_{ij}
\]
where \( d_{ij} \) is the distance between area \( i \) and site \( j \).

\paragraph{Constraints}
\begin{enumerate}
    \item Site Selection Limit: Only four sites can be selected.
    \[
    \sum_{j=1}^{10} X_j = 4
    \]
    \item Assignment to Selected Sites: Residents can only be assigned to selected sites.
    \[
    Y_{ij} \leq X_j \quad \text{for all} \quad i, j
    \]
    \item Capacity Constraint: The number of residents assigned to a clinic must not exceed its capacity.
    \[
    \sum_{i=1}^{50} r_i \times Y_{ij} \leq C_j \times X_j \quad \text{for all} \quad j
    \]
    \item Assignment of Each Area: Each residential area must be assigned to exactly one clinic.
    \[
    \sum_{j=1}^{10} Y_{ij} = 1 \quad \text{for all} \quad i
    \]
    \item Binary and Integer Constraints:
    \begin{itemize}
        \item \( X_j \in \{0, 1\} \) for all \( j \).
        \item \( Y_{ij} \in \{0, 1\} \) for all \( i, j \).
    \end{itemize}
\end{enumerate}

This model aims to determine the optimal selection and assignment of vaccination sites to minimize the total travel distance for residents under the given constraints and assumptions.


\subsection{Radiotherapy Treatment}

We are delighted to welcome you, our newest intern on the Analytics team of Massachusetts General Hospital! You have been placed in a challenging role where you will be tasked with solving a real-world problem in the field of medical physics. We are building a pilot program in Boston, and if successful, your work could be applied widely in hospitals with limited capacity in many countries.

You are responsible for determining the best treatment plan for 17 patients who require radiotherapy. Your goal is to optimize the use of two possible treatments: photon therapy and proton therapy. While proton therapy is known to target tumors more precisely, it is also more expensive and has limited capacity in many countries. Therefore, you will need to balance the benefits of proton therapy with its limitations and cost to create an effective treatment plan for each patient.

To determine the best course of action for each patient, you will use a scoring system called the Biological Equivalent Dose (BED). This system allows you to calculate the effectiveness of each patient's treatment plan by considering the number of proton fractions that can be used while still achieving the highest possible BED.

We have $n=17$ patients who need radiotherapy. Each patient $i$ needs 15 fractions, which can be photon fractions, proton fractions, or a mix of photon and proton fractions (e.g., 4 proton fractions and 11 photon fractions).  We want to use the limited proton therapy capacity as best as possible. We can calculate the BED score for each patient when $p$ proton fractions and $15-p$ photon fractions are used, as BED$_i$($p$,$15-p$), i.e., the BED when $p$ proton and $15-p$ photon fractions are delivered for patient $i$. The higher the score, the better.

The data file ``outcome\_data.csv" on Canvas contains a matrix of $17$ rows and $16$ columns. In particular, the number at the $(i,j)$ position is BED$_{i}$($j-1$,$15-j+1$)\footnote{Notice Julia starts indexing at 1, hence this $j-1$, $15-j+1$ as an adjustment.}.

Suppose that the total maximal capacity is $C$ proton fractions. To maximize the total BED scores for all the patients, which patients should get proton fractions, and how many should they get? Formulate an integer linear optimization model to solve this problem. Assume you know the value BED$_i$($j$,$15-j$) for each patient $i \in [1,17].$

\subsubsection{Correct Answer from ChatGPT}

To formulate an integer linear optimization model for this problem, we define the decision variables, the objective function, and the constraints. The goal is to maximize the total Biological Equivalent Dose (BED) scores for all patients, considering the limited capacity of proton fractions.

\paragraph*{Decision Variables}
Define a matrix \( x_{i,p} \) where:
\begin{itemize}
    \item \( i \) is the patient index (\( i \in \{1, 2, ..., 17\} \)).
    \item \( p \) is the number of proton fractions (\( p \in \{0, 1, ..., 15\} \)).
    \item \( x_{i,p} \) is a binary variable that takes the value 1 if patient \( i \) receives \( p \) proton fractions (and \( 15-p \) photon fractions), and 0 otherwise.
\end{itemize}

\paragraph*{Objective Function}
The objective is to maximize the total BED across all patients. Therefore, the objective function is:
\[
\text{Maximize } \sum_{i=1}^{17} \sum_{p=0}^{15} \text{BED}_i(p, 15-p) \times x_{i,p}
\]

\paragraph*{Constraints}
\begin{enumerate}
    \item \textbf{Capacity Constraint}: The total number of proton fractions used across all patients must not exceed the total capacity \( C \).
    \[
    \sum_{i=1}^{17} \sum_{p=0}^{15} p \times x_{i,p} \leq C
    \]

    \item \textbf{Treatment Constraint}: Each patient should receive exactly one combination of proton and photon fractions.
    \[
    \sum_{p=0}^{15} x_{i,p} = 1 \quad \text{for each patient } i
    \]

    \item \textbf{Integer Constraint}: The decision variables \( x_{i,p} \) are binary.
    \[
    x_{i,p} \in \{0, 1\}
    \]
\end{enumerate}

\paragraph*{Model Summary}
\begin{itemize}
    \item \textbf{Objective}: Maximize \( \sum_{i=1}^{17} \sum_{p=0}^{15} \text{BED}_i(p, 15-p) \times x_{i,p} \)
    \item \textbf{Subject to}:
    \begin{itemize}
        \item Capacity Constraint: \( \sum_{i=1}^{17} \sum_{p=0}^{15} p \times x_{i,p} \leq C \)
        \item Treatment Constraint: \( \sum_{p=0}^{15} x_{i,p} = 1 \) for each \( i \)
        \item Integer Constraint: \( x_{i,p} \in \{0, 1\} \)
    \end{itemize}
\end{itemize}

This model can be solved using integer linear optimization techniques to determine the optimal distribution of proton and photon fractions for each patient, given the total capacity \( C \) of proton fractions.

\subsubsection{Wrong answer from ChatGPT}

Note: The model generated a nonlinear objective function using the term \textbf{Maximize} \(\sum_{i=1}^{17} \text{BED}_i(x_{i}, 15 - x_{i})\). However, the correct formulation should have included the term \textbf{Maximize} \(\sum_{i=1}^{17} \sum_{p=0}^{15} \text{BED}_i(p, 15-p) \times x_{i,p}\), as demonstrated above. This represents a subtle error that can easily go unnoticed.

Indeed, both the incorrect and correct terms involve summations and the BED function, which could lead to a quick glance overlooking the additional summation and variable in the correct term. Without a deep understanding of the context or the specific requirements of the model, one might not realize that the additional summation and the $x_{i,p}$ term are necessary to obtain an integer linear optimization model.

When prompted that the formulation

ChatGPT's answer:

To formulate an integer linear optimization model for this problem, we need to consider the following elements:

\begin{itemize}
    \item \textbf{Decision Variables:} Represent the number of proton fractions allocated to each patient.
    \item \textbf{Objective Function:} Maximize the total Biological Equivalent Dose (BED) score across all patients.
    \item \textbf{Constraints:} Include the total capacity of proton therapy and the requirement for each patient to receive 15 fractions in total.
\end{itemize}

\paragraph*{Decision Variables}
Let \( x_{i} \) be the number of proton fractions allocated to patient \( i \), where \( i \) ranges from 1 to 17. \( x_{i} \) is an integer.

\paragraph*{Objective Function}
\textbf{Maximize} \( \sum_{i=1}^{17} \text{BED}_i(x_{i}, 15 - x_{i}) \)
This function sums the BED scores for all patients, maximizing the total effectiveness of the treatments.

\paragraph*{Constraints}
\begin{enumerate}
    \item \textbf{Proton Capacity Constraint:} \( \sum_{i=1}^{17} x_{i} \leq C \)
    The total number of proton fractions used should not exceed the maximum capacity \( C \).

    \item \textbf{Fraction Constraint for Each Patient:} \( 0 \leq x_{i} \leq 15 \) for all \( i \in [1,17] \)
    Each patient must receive a total of 15 fractions, which can be a mix of proton and photon fractions.

    \item \textbf{Integrality Constraint:} \( x_{i} \) are integers.
\end{enumerate}

\paragraph*{Integer Linear Optimization Model}
The final model can be represented as follows:

\textbf{Maximize:} \( \sum_{i=1}^{17} \text{BED}_i(x_{i}, 15 - x_{i}) \)

\textbf{Subject to:}
\begin{enumerate}
    \item \( \sum_{i=1}^{17} x_{i} \leq C \)
    \item \( 0 \leq x_{i} \leq 15 \) for all \( i \in [1,17] \)
    \item \( x_{i} \) are integers.
\end{enumerate}

This model will allocate the proton therapy fractions to the patients in a way that maximizes the overall effectiveness of the treatments, considering the constraints of the total proton capacity and individual patient needs.

\section{Human Crafted Optimization Models For Example \ref{exa:Mobian} \label{apx:mobian-optimization-models} }

In this section, we give two hub location models developed for  by an Optimization expert (without using ChatGPT) for the problem of Example \ref{exa:Mobian}. The two models are equivalent. However, the second model is computationally more tractable since it contains fewer variables. 
We first introduce sets, parameters, and variables that we use in the models.
\\

\noindent
{\bf Sets:} \\
\begin{align*}
    H &: \text{Set of hub indexes, where $h=1,...,N$ refers to all existing hubs and} &&&&&\\
  &  \text{    indexes $h=N+1,...,M$ corresponds to all potential hub locations.} &&&&&\\
   P &: \text{Index set of POIs (size $n$).}&&&&&\\
    S &: \text{Index set of all junction roads (size $q$).}&&&&&\\
    F &= \{s\in S, h\in H, p\in P \mid c_{sp}+\Delta < c_{sh}+b_{hp} \text{ or } b_{hp}> T \text{ or } d_{hp}<D \text{ or } d_{sp}-d_{hp}<\tau\} &&&&&
\end{align*}
\\

\noindent
{\bf Parameters:}
\begin{align*}
v_{sp} &: \text{Demand for POI $p$, coming from junction $s$}\\
c_{sp} &: \text{Car travel time from junction $s$ to POI $p$ (min)}\\
c_{sh} &: \text{Car travel time from junction $s$ to hub $h$ (min)}\\
b_{hp} &: \text{Bike travel time from hub $h$ to POI $p$ (min)}\\
d_{sp} &: \text{Distance from junction $s$ to POI $p$ (meters) }\\
d_{hp} &: \text{Distance from hub $h$ to POI $p$ (meters) }\\
T &: \text{Longest accepted bike travel time from a hub to a POI (min)}\\
N &: \text{Number of existing hub locations}\\
U &: \text{Maximum number of new hubs planned to be opened}\\
D &: \text{Minimum distance between a hub and a POI (meters)}\\
\Delta &: \text{Maximum allowed extra travel time by bike from car travel time (min)}\\
\tau &: \text{Distance separating a hub lopcation from a POI (meters)}
\end{align*}
\\

\noindent
{\bf Variables:}
\begin{align*}
    x_{shp} & = \begin{cases} 1 \text{ if hub $h$ is assigned to POI $p$ from junction $s$ and} \\
     \text{  \hspace{5mm}   $c_{sp}+\Delta > c_{sh} + b_{hp}$, $d_{sp}-d_{hp}>\tau$, $b_{hp} < T$ and $d_{hp} > D$}\\
    0 \text{ if not}  \end{cases} \\
    y_h & = \begin{cases} 1 \text{ if hub $h$ is opened}\\ 0 \text{ else} \end{cases}
\end{align*}
\\

\noindent {\bf Model 1} \\

\begin{align*}
\text{max} \quad & \sum_{s \in S} \sum_{h \in H} \sum_{p \in P} v_{sp} x_{shp} \\
\text{s.t.} \quad & \sum_{h=N+1}^{M} y_h \leq U & & \quad \text{Bound number of opened hubs} \\
& \sum_{h=1}^{N} y_h = N & & \quad \text{Set the first $N$ hubs to be open} \\
& \sum_{h=1}^{M} x_{shp} \leq 1 & \forall s \in S, p \in P & \quad \text{Ensure every OD-pair travels via at most one hub} \\
& x_{shp} \leq y_h & \forall s \in S, p \in P, h \in H & \quad \text{An OD-pair can only travel via an opened hub} \\
& x_{shp} = 0 & \forall s, h, p \in F & \quad \text{All OD-pairs that are in the 'not allowed' set $F$ are set to 0} \\
& x_{shp}, y_h \in \{0, 1\} & \forall s \in S, h \in H, p \in P & \quad \text{Domain constraints}
\end{align*}

The objective is to maximize the covered demand. The first constraint is an upper bound on the number of opened hubs. The second constraint is to set the first $m$ existing hubs to be open. 
The third constraint is to ensure that every junction-POI pair can be assigned to at most one hub location.
The fourth constraint is to ensure that every junction-POI pair can only travel via an opened hub. The fifth constraint is to ensure that all junction-POI pairs that are in the 'not allowed' set $F$ are set to $0$. The last constraint ensures that the variables are binary.  
\\

\noindent
Only one other set and variable are needed for the second model.  
\\

\noindent
{\bf Set:}
\begin{align*}
    \delta_h &= \{(s,p)\in S\times P \mid c_{sp}+\Delta \geq c_{sh}+b_{hp} \text{ and } b_{hp}\leq T \text{ and } d_{hp}
    \geq D \text{ and } d_{sp}-d_{hp}\geq\tau\}. &&&&&
\end{align*}
\\

\noindent {\bf Variable:}
\begin{align*}
    x_{sp} & = \begin{cases} 1 \text{ if $\exists$ an opened hub $h\in H$ such that} \\
    \text{    \hspace{5mm} $c_{sp}+\Delta \geq c_{sh} + b_{hp}$, $d_{sp}-d_{hp}>\tau$, $b_{hp} \leq T$ and $d_{hp} > D$}\\
    0 \text{ otherwise.}  \end{cases} 
\end{align*}
\\

\noindent{\bf Model 2}

\begin{align*}
\text{max} \quad & \sum_{s \in S} \sum_{p \in P} v_{sp} x_{sp} \\
\text{s.t.} \quad & x_{sp} \leq \sum_{h \in H:(s,p) \in \delta_h} y_h & \forall s \in S, p \in P & \quad \text{Traveling from junction $s$ to POI $p$ is only possible if one of the nearby hubs is opened} \\
& \sum_{h=N+1}^{M} y_h \leq U & & \quad \text{No more than $U$ locations can be opened} \\
& \sum_{h=1}^{N} y_h = N & & \quad \text{Existing hubs are fixed to be opened} \\
& x_{sp}, y_h \in \{0, 1\} & \forall s \in S, h \in H, p \in P & \quad \text{Domain constraints}
\end{align*}


Only the first constraint is new: this constraint ensures that the junction-POI pair $(s,p)$ is `covered' if one of the nearby hubs is opened.

\section{Experiments with Gap 3}

\subsection{Answer from ChatGPT to reduce the number of variables \label{apx:req3-exp1}}

The following is a correct answer from ChatGPT that was given the use case from Example \ref{exa:Mobian} as before, to which we appended Model 1 from Section \ref{sec:exp-req2}a nd tasked to: ``Improve this formulation by reducing the number of variables and making it more efficient. The formulation should be equivalent, but you must reduce the number of variables by modifying the objective and constraints. Explain the reasoning first and then detail the equivalent reformulation.''
For brevity, we only include the new formulation given by ChatGPT below.

\subsubsection*{Revised Sets, Parameters, and Variables}

\textbf{Sets and Parameters:} (Remain the same as in the original formulation)

\textbf{Variables:}
\begin{itemize}
    \item \( y_h \): Binary variable. 1 if hub \( h \) is opened; 0 otherwise.
    \item \( z_{sp} \): Binary variable. 1 if Point of Interest (POI) \( p \) is assigned to any hub from junction \( s \) and satisfies the defined constraints; 0 otherwise.
\end{itemize}

\subsubsection*{Objective and Constraints}

\textbf{Objective:}
\begin{itemize}
    \item Maximize the total demand for POIs, considering the assignment to hubs:
    \[ \text{Maximize} \quad \sum_{s \in S} \sum_{p \in P} v_{sp} z_{sp} \]
\end{itemize}

\textbf{Constraints:}
\begin{enumerate}
    \item Bound the number of opened hubs:
    \[ \sum_{h=N+1}^{M} y_h \leq U \]
    (Ensure the number of new hubs opened does not exceed the planned maximum.)

    \item Set the existing hubs to be open:
    \[ \sum_{h=1}^{N} y_h = N \]

    \item Assign POI to a suitable hub:
    \[ z_{sp} \leq \sum_{h=1}^{M} y_h \times \mathbb{I}_{\{h \text{ satisfies conditions for } p \text{ from } s\}} \quad \forall s \in S, p \in P \]
    (Here, \( \mathbb{I} \) is an indicator function that checks if hub \( h \) satisfies the necessary conditions for POI \( p \) from junction \( s \).)

    \item Domain constraints for the binary variables:
    \[ z_{sp}, y_h \in \{0, 1\} \quad \forall s \in S, p \in P, h \in H \]
\end{enumerate}

\subsubsection*{Explanation of the Reformulation}
\begin{itemize}
    \item This new formulation simplifies the model by reducing the number of binary variables. The complex variable \( x_{shp} \) from the original formulation is removed and replaced with \( z_{sp} \).
    \item The objective function is now more straightforward. It directly maximizes the demand coverage with a simpler summation structure.
    \item Constraints are adjusted to incorporate the logic that was previously captured by \( x_{shp} \), using an indicator function within the constraints.
    \item Overall, this reformulation aims to maintain the decision-making capability of the model while enhancing its computational efficiency by simplifying its structure.
\end{itemize}

\subsection{Wrong formulation from ChatGPT \label{apx:gap3-incorrect1}}

The model suggested by ChatGPT below is not correct. It contains a max operator and variables that indicate an index. ChatGPT’s model is also not a mixed-integer linear optimization model.

\subsubsection*{Sets and Parameters (Unchanged)}
\begin{itemize}
  \item Sets: \( H, P, S, F \) as defined in the original model.
  \item Parameters: \( v_{sp}, c_{sp}, c_{sh}, b_{hp}, d_{sp}, d_{hp}, T, N, U, D, \Delta, \tau \) as defined in the original model.
\end{itemize}

\subsubsection*{Modified Variables}
\begin{itemize}
  \item \( y_h \): Unchanged, indicating if hub \( h \) is opened.
  \item \( z_{sp} \): Indicates the hub assigned to POI \( p \) from junction \( s \). This variable takes a value from the set of possible hubs \( H \).
\end{itemize}

\subsubsection*{Objective Function}
Maximize 
\[
\sum_{s \in S} \sum_{p \in P} v_{sp} \times \text{score}(z_{sp}, s, p)
\]
Where \(\text{score}(z_{sp}, s, p)\) is a function that returns 1 if the hub \( z_{sp} \) satisfies the conditions for the OD-pair \((s, p)\) (i.e., \( c_{sp}+\Delta > c_{sh} + b_{hp} \), \( d_{sp}-d_{hp}>\tau \), \( b_{hp} < T \), \( d_{hp} > D \)), and 0 otherwise.

\subsubsection*{Constraints}
\begin{enumerate}
  \item \( \sum_{h=N+1}^{M} y_h \leq U \)
    \begin{itemize}
      \item Bound the number of opened hubs.
    \end{itemize}
  \item \( \sum_{h=1}^{N} y_h = N \)
    \begin{itemize}
      \item Set the first \( N \) hubs to be open.
    \end{itemize}
  \item \( z_{sp} \leq M \times \max(y_h) \) for all \( s \in S, p \in P \)
    \begin{itemize}
      \item An OD-pair can only travel via an opened hub.
    \end{itemize}
  \item \( z_{sp} \) is not assigned to any hub in the 'not allowed' set \( F \) for all \( s, p \).
  \item Domain constraints: \( y_h \in \{0, 1\} \) for all \( h \in H \); \( z_{sp} \in H \) for all \( s \in S, p \in P \).
\end{enumerate}

\subsection{Wrong answer from ChatGPT}

ChatGPT uses the indices hub and POI for the variables, which is not correct.

\subsubsection*{Revised Sets and Parameters}
Keeping the sets and parameters largely unchanged, but we introduce a new parameter to reduce variables:
\begin{align*}
    H, P, S & \quad \text{(as defined earlier)} \\
    F & \quad \text{(as defined earlier)} \\
    v_{p} & \quad \text{Aggregated demand for POI $p$ from all junctions}
\end{align*}

\subsubsection*{Revised Variables}
\begin{align*}
    z_{hp} & = \begin{cases} 
    1 & \text{if hub $h$ is assigned to POI $p$ and meets all constraints} \\
    0 & \text{if not} 
    \end{cases} \\
    y_h & \quad \text{(as defined earlier)}
\end{align*}

\subsubsection*{Revised Objective and Constraints}
\begin{align*}
    \text{max} \quad & \sum_{h \in H} \sum_{p \in P} v_{p} z_{hp} \\
    \text{s.t.} \quad & \sum_{h=N+1}^{M} y_h \leq U & & \quad \text{Bound number of opened hubs} \\
    & \sum_{h=1}^{N} y_h = N & & \quad \text{Set the first $N$ hubs to be open} \\
    & \sum_{h=1}^{M} z_{hp} \leq 1 & \forall p \in P & \quad \text{Ensure every POI is served by at most one hub} \\
    & z_{hp} \leq y_h & \forall p \in P, h \in H & \quad \text{A POI can only be served by an opened hub} \\
    & z_{hp} = 0 & \forall h, p \in F' & \quad \text{All POI-hub pairs that are in the 'not allowed' set $F'$ are set to 0} \\
    & z_{hp}, y_h \in \{0, 1\} & \forall h \in H, p \in P & \quad \text{Domain constraints}
\end{align*}

\subsection{Wrong Formulation from ChatGPT}

This model suggested by ChatGPT is not correct. The conditions are just put as constraints without variables.

\subsubsection*{Sets:}
\begin{align*}
    H &: \text{Set of hub indexes, where } h=1,...,N \text{ refers to all existing hubs and} \\
       & \text{indexes } h=N+1,...,M \text{ correspond to all potential hub locations.} \\
    P &: \text{Index set of POIs (Points of Interest), size } n. \\
    S &: \text{Index set of all junction roads, size } q.
\end{align*}

\subsubsection*{New Parameters:}
\begin{align*}
    \Delta &: \text{Maximum allowed extra travel time by bike from car travel time (min)}.
\end{align*}

\subsubsection*{Variables:}
\begin{align*}
    y_h & = \begin{cases} 1 \text{ if hub } h \text{ is opened,} \\ 0 \text{ otherwise} \end{cases} \\
    z_{sp} & = \begin{cases} h \text{ if POI } p \text{ from junction } s \text{ is served by hub } h, \\ 0 \text{ if POI } p \text{ from junction } s \text{ is not served by any hub} \end{cases}
\end{align*}

\subsubsection*{Objective Function and Constraints:}

Objective:
\begin{equation*}
    \text{Maximize} \quad \sum_{s \in S} \sum_{p \in P} v_{sp} \cdot (z_{sp} > 0)
\end{equation*}

Constraints:
\begin{align*}
    \sum_{h=N+1}^{M} y_h &\leq U & & \text{(Bound number of opened hubs)} \\
    \sum_{h=1}^{N} y_h &= N & & \text{(Set the first } N \text{ hubs to be open)} \\
    z_{sp} &\leq M \cdot y_h & \forall s \in S, p \in P, h \in H & \text{(A POI can only be served by an opened hub)} \\
    c_{sp} + \Delta &\geq c_{sh} + b_{hp} & \forall s \in S, p \in P, h \in H & \text{(Ensure travel time constraints)} \\
    d_{hp} &\geq D & \forall h \in H, p \in P & \text{(Minimum distance between a hub and a POI)} \\
    y_h, z_{sp} &\in \{0, 1, ..., M\} & \forall s \in S, h \in H, p \in P & \text{(Domain constraints)}
\end{align*}

\end{document}